\documentclass[sigconf,]{acmart}

\copyrightyear{2024}
\acmYear{2024}
\setcopyright{acmlicensed}\acmConference[CAIN 2024]{Conference on AI Engineering Software Engineering for AI}{April 14--15, 2024}{Lisbon, Portugal}
\acmBooktitle{Conference on AI Engineering Software Engineering for AI (CAIN 2024), April 14--15, 2024, Lisbon, Portugal}
\acmDOI{10.1145/3644815.3644962}
\acmISBN{979-8-4007-0591-5/24/04}

\usepackage{graphicx}
\usepackage{multirow}
\usepackage{xspace}
\usepackage{tcolorbox}
\usepackage{url}

\newcommand{\etal}{\emph{et~al.}\xspace}

\title{Identifying architectural  design decisions for achieving green ML serving}

\author{Francisco Durán}
\email{francisco.duran.lopez@upc.edu}
\affiliation{%
  \institution{Universitat Politècnica de Catalunya}
  \city{Barcelona}
  \state{Catalunya}
  \country{Spain}
}

\author{Silverio Martínez-Fernández}
\email{silverio.martinez@upc.edu}
\affiliation{%
  \institution{Universitat Politècnica de Catalunya}
  \city{Barcelona}
  \state{Catalunya}
  \country{Spain}
}

\author{Matias Martinez}
\email{matias.martinez@upc.edu}
\affiliation{%
  \institution{Universitat Politècnica de Catalunya}
  \city{Barcelona}
  \state{Catalunya}
  \country{Spain}
}

\author{Patricia Lago}
\email{p.lago@vu.nl}
\affiliation{%
  \institution{Vrije Universiteit Amsterdam}
  \city{Amsterdam}
  \country{Netherlands}
}

\twocolumn
\begin{document}

\begin{abstract}


The growing use of large machine learning models highlights concerns about their increasing computational demands. While the energy consumption of their training phase has received attention, fewer works have considered the inference phase. For ML inference, the binding of ML models to the ML system for user access, known as \textit{ML serving}, is a critical yet understudied step for achieving efficiency in ML applications.

We examine the literature in ML architectural design decisions and Green AI, with a special focus on ML serving. The aim is to analyze ML serving architectural design decisions for the purpose of understanding and identifying them with respect to quality characteristics from the point of view of researchers and practitioners in the context of ML serving literature.

Our results \textit{(i)} identify ML serving architectural design decisions along with their corresponding components and associated technological stack, and \textit{(ii)} provide an overview of the quality characteristics studied in the literature, including energy efficiency.

This preliminary study is the first step in our goal to achieve green ML serving. Our analysis may aid ML researchers and practitioners in making green-aware architecture design decisions when serving their models. 
\end{abstract}

\maketitle

\section{Introduction}
Due to the increasing use of large Machine Learning (ML) models, the required computational resources for both training (i.e. in which the model is built) and inference (i.e. in which the model is used to make predictions from new data) phases are increasing every year \cite{lwakatare2019taxonomy,Schwartz2020GreenAI}. This has raised several concerns including financial and environmental costs \cite{Strubell2019Monitoring} since the majority of ML practitioners are focused on improving the models w.r.t. accuracy. 

In the inference phase, the trained models are bound into the ML application to allow users to make predictions. This process is also known as \textit{ML serving} \cite{zhang2020inferbench, kumara2022requirements}. The significance of ML serving architectural design decisions (ADDs) cannot be overstated, as they can play an important role in determining their efficiency and sustainability \cite{stier2015model}. 
Software engineering and software architecture approaches to ML are still emerging compared to well-established traditional software systems architecture patterns and design decisions. In that way, practitioners may have difficulties in identifying the most efficient available choices, particularly when it comes to making specific decisions~\cite{warnett2022architectural}.
The main decision practitioners must make is how to integrate an ML model in an application~\cite{kumara2022requirements,warnett2022architectural}, having at least three alternatives: \textit{(i) model in monolith, (ii) model as a service, or (iii) no integration (results are used and there is no request from user e.g. forecast)}. In this paper, we focus on ML inference when ML models are deployed as a service.

In the development of traditional software systems, architectural decision-making is a primary approach to improve different software quality characteristics, including sustainability \cite{ISO25010,stier2015model, calero2015green}. Nonetheless, there is evidence suggesting that current architecture-centric methods do not adequately consider the potential for optimizing ML-based systems for efficiency \cite{Schwartz2020GreenAI, wu2022sustainable}.
In this study, we made an overview of the quality characteristics studied in the selected literature, including energy efficiency\cite{calero2015green}.

This research short paper presents a preliminary study of green ML serving ADDs and their studied quality characteristics. The paper aims at identifying the main ML serving ADDs from literature, conducting a comparative analysis focusing on their components, quality characteristics, and trade-offs, and discussing research opportunities in the area of ML serving strategies and Green AI.

The paper is organized as follows. In Section 2, we provide an introduction to the domain of our research and pertinent prior research. Moving to Section 3, we present the research questions and the study design. Subsequently, Sections 4 and 5 present our results and discuss the main threats to validity, respectively. Finally, Section 6 draws discussions, conclusions and directions for future work.

\section{Background and Related Work}

 There is scarce research focused on how to deploy ML models more efficiently once you have a trained model \cite{desislavov2023trends,verdecchia2023systematic}. Consequently, there is a scarcity of literature on ML serving and its design decisions in this context. 

Zhang \emph{et al.} \cite{zhang2018} evaluated combinations of hardware and software (ML frameworks) packages when running inferences of CNN models on edge devices regarding latency, memory footprint and energy. Georgiou \emph{et al.} \cite{georgiou2022green} compared the energy consumption and run-time performance of two of the most popular DL frameworks (TensorFlow and PyTorch) during both phases, training and inference. Koubaa \emph{et al.} \cite{koubaa2021cloud} analyzed three deployment strategies (cloud, edge and using both) and two optimized model formats (TensorRT and TFLite). Hampau \emph{et al.} \cite{hampau2022empirical} studied the impact on energy consumption and performance of using different containerization strategies in only computer vision tasks on edge devices.  Klimiato \emph{et al.} \cite{klimiato2022utilizing} analyzed deployment solutions from only one type of ML serving infrastructure (DL-specific software) w.r.t. energy-related and performance-related measurements in only computer vision task such as image classification. Lenherr \emph{et al.} \cite{lenherr2021new} proposed sustainability metrics for edge training and inference. They used three different runtime engines and model formats to run their experiments.

In the benchmark of Zhang \emph{et al.} \cite{zhang2020inferbench}, they studied both DL-specific software and runtime engines, regarding GPU usage and some fine-grained characteristics such as batch tail latency, batch size and dynamic batching. 

Furthermore, Escribano \emph{et al.} \cite{escribano2023energy} focused his work on the energy consumption of free plans cloud computing from different cloud providers, using NLP models and did not compared the ML serving infrastructure. Yao \etal \cite{yao2021evaluating} conducted an empirical study on model-level and layer-level energy efficiency of CNN models to find the most efficient configuration for CNN inference on high-performance GPUs. Yarally \etal \cite{yarally2023batching} explored the effect of input batching on the energy consumption and response times of five DL models finding that batching can significantly reduce both metrics.

Regarding the design decisions when serving model, Warnett \emph{et al.} \cite{warnett2022architectural} identified design decisions for model deployment or inference phase in general, as they identified decisions considering not only serving but other pipelines or steps (e.g., monitoring, testing, versioning, etc.). Kumara \etal \cite{kumara2022requirements} studied the requirements and components of a reference architecture for MLOps but considered ML serving just at a high level. For their part, Lwakatare \etal \cite{lwakatare2019taxonomy} identified and classified software engineering challenges from different companies when developing ML systems, developing a taxonomy of prevalent issues. Franch \etal \cite{franch2022architectural} proposed a preliminary ontology for architectural decision making in the context of AI, highlighting the impact of these decisions on architectural elements and their influence on quality attributes. Morevover, recent benchmarks for ML models, evaluated the different operations within ML inference systems \cite{reddi2020mlperf, zhang2020inferbench}.



In contrast to previous studies, this preliminary study aims to identify various ML serving options, and analyze their quality aspects. This will guide  ML practitioners into the green-aware architectural design of ML serving systems.

\section{Research methodology}

\subsection{Goal, research questions and hypothesis}

The research goal \cite{basili1994goal} of this study is to analyze \textit{ML serving ADDs} for the purpose of \textit{understanding and identifying them} with respect to \textit{quality characteristics} from the point of view of \textit{researchers and practitioners} in the context of \textit{ML serving literature}.

\begin{itemize}
\item [RQ1:] What are the ADDs of ML serving?
\item [RQ2:] What are the studied quality characteristics of those ADDs?
\end{itemize}

\subsection{Literature search and selection}
For the literature search, we adopted a seed papers and snowballing strategy. It is important to note that this study is a preliminary exploration, and the methodology outlined here may evolve in subsequent phases of research.
Despite these limitations, we believe that the literature search presented in this paper provides a valuable starting point for understanding the current state of research on ADDs for green ML serving. First, we initiated our research by selecting as seed papers the peer-reviewed primary studies of two recent reviews. 
The first review focuses on ADDs for ML deployment, exploring the critical considerations that underlie the entire process of deployment (inference phase) \cite{warnett2022architectural,warnett2022dataset}. The second review a systematic Green AI literature review from 2015 to 2022 \cite{verdecchia2023systematic}. This review provides an overview of the landscape of Green AI and characterizes it, including ML deployment and ADD aspects. With this, we initially selected 6 papers as seed papers. We include papers that consider \textit{(i)} inference phase and \textit{(ii)} the report of deployment strategy and/or used software technological stack in inference phase.

In order to extend the search, we employed a  forward snowballing using Google Scholar. Thus, we respectively chose 6 more papers that fulfill our selection criteria. This method allowed us to consider 12 (from seed and snowballing  papers) works that further expanded on the green AI-related aspects of ML serving or ADDs of ML serving.

\subsection{Data collection and analysis}

We defined the properties of data that address our research questions including bibliometrics, ML serving ADDs, and quality characteristics. Each study is
carefully analyzed to extract this data into a spreadsheet. The replication package includes the extracted data, which is available online \footnote{Please refer to \url{https://zenodo.org/doi/10.5281/zenodo.10547444}}.

After data extraction, we performed data analysis. 
For responding RQ1, we conduct a thematic analysis as proposed in \cite{cruzes2011recommended} for: analyzing  the ML serving ADDs employed in the field, identifying the design decisions, and performing a classification of them based on their direct impact on the ML serving process. Additionally, a detailed exploration of the architecture of identified design decisions was undertaken, emphasizing their architectural components and interrelations.

For responding RQ2, based on the ISO 25010 \cite{ISO25010} and the extended quality model with the `\textit{greenability}' quality characteristic \cite{calero2015green}, we explore the quality characteristics associated with each design decision, previously studied in literature.
We identified 8 different quality characteristics in the selected literature:

\begin{itemize}
    \item Energy efficiency refers to the degree of efficiency with which a software product consumes energy when performing its functions \cite{calero2015green}.
    \item Performance efficiency refers to the performance relative to the amount of resources used under stated conditions  \cite{ISO25010}. 
    \item Maintainability refers to the degree of effectiveness and efficiency with which a product or system can be modified to improve it, correct it or adapt it to changes in environment, and in requirements \cite{ISO25010}.
    \item Analysability refers to the degree of effectiveness and efficiency with which it is possible to assess the impact on a product or system of an intended change to one or more of its parts, or to diagnose a product for deficiencies or causes of failures, or to identify parts to be modified \cite{ISO25010}.
    \item Usability refers to the degree to which a product or system can be used by specified users to achieve specified goals with effectiveness, efficiency and satisfaction in a specified context of use \cite{ISO25010}. 
    \item Scalability refers to the ease with which an application or component can be modified to expand its existing capabilities \cite{miguel2014review}.
    \item Portability refers to the degree of effectiveness and efficiency with which a system, product or component can be transferred from one hardware, software or other operational or usage environment to another\cite{ISO25010}.
    \item Interoperability refers to the degree to which two or more systems, products or components can exchange information and use the information that has been exchanged \cite{ISO25010}.
\end{itemize}

\section{Results}

\begin{figure*}[htp]
    \centering
    \includegraphics[width=0.71\textwidth]{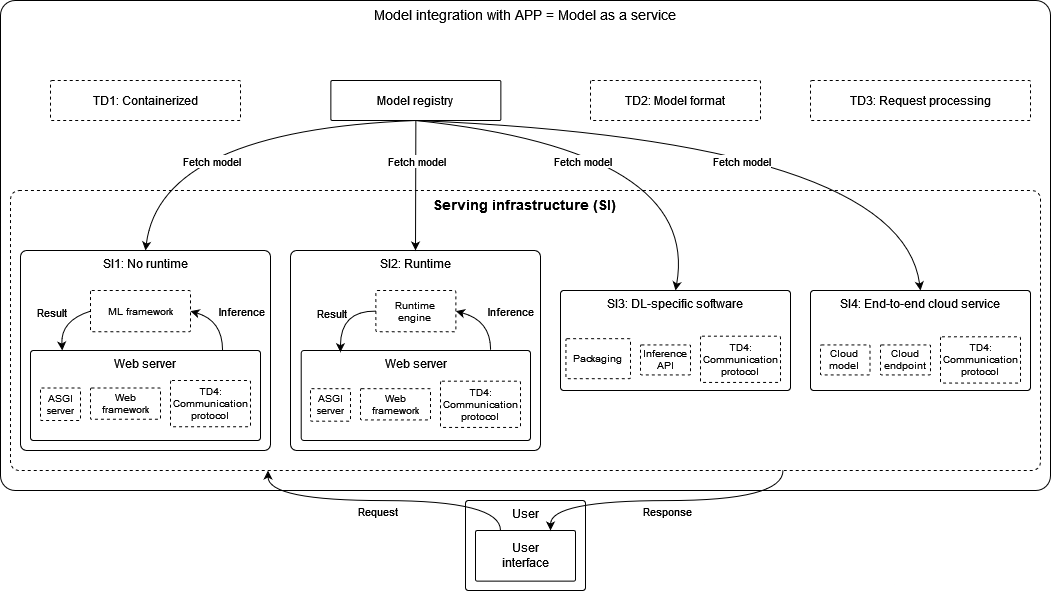}
    \caption{ML serving ADDs.}
    \label{fig:serving_infr}
\end{figure*}

In the following, we present our results and the way they address RQ1 and RQ2. To do so, we refer to Figure~\ref{fig:serving_infr} and Table~\ref{tab:table_decisions}.

\subsection{What are the architectural design decisions of ML serving? (RQ1)}

When the ML models are integrated to the ML application in a model as a service manner, we found one principal design decision, which is the \textit{Serving Infrastructure (SI)}, in which practitioners have mainly four different options to serve their models. The results are presented in the Table ~\ref{tab:table_decisions}.

The four identified serving infrastructure (\textit{SI1, SI2, SI3} and \textit{SI4}) architectures are presented in Figure \ref{fig:serving_infr} and a description of them is presented in the following paragraphs:

\begin{itemize}
    \item `\textit{No runtime engine and web server}' \textit{(SI1)}: This option refers to use the ML framework (e.g., TensorFlow, and PyTorch), and a web framework and an ASGI server to build the API in order to provide users access to make inferences. This is the simplest choice and it is traditionally used for training, but can be used for inference. The operation mechanism is to load the model from model registry, when user sends a request to the web server it loads the model using the ML framework, runs inference, and sends back the response with the prediction to the user.
    For instance, this configuration involves an API, the server receives the input, redirects it to the ML framework to process the input using the model, obtains the result, and then returns the response to the user (see figure 2 from \cite{koubaa2021cloud}).
    \item `\textit{Runtime engine}' \textit{(SI2)}: This option is very similar to the previous, with the difference that the `\textit{Runtime engine}' (e.g., ONNX runtime, torch.jit, OpenVINO) is used to make the inference instead of just using the ML framework. A `\textit{Runtime engine}' is a software that applies optimizations to the model, such as graph optimizations or just-in-time compilation.
    The operational mechanism involves loading the model from the model registry. Then, when a user sends a request to the web server, the model is loaded using the `\textit{Runtime engine}', runs inference, and sends back the response with the prediction to the user.
    The optimized ONNX runtime is employed to load ONNX models and wrap them as a service using a web framework such as FastAPI. Once the service is running, the user can make predictions through the service, using a communication protocol like gRPC (see figure 6 from \cite{zhang2020inferbench}).
    
    
    \item `\textit{DL-specific software}' \textit{(SI3)}: This option refers to use the serving software created for the ML framework, specifically for a production-like serving. The operation mechanism is to package the model(s) into the `\textit{DL-specific software}' according to the provided handlers or provided configuration from developer, after that the software configures endpoints through an Inference API. Unlike the two previous infrastructures, here it is not necessary to manually build an API, when user sends a request to the `\textit{DL-specific software}' it runs the inference using the configuration provided in the packaging of the models, and sends back the response with the prediction to the user. 

    \item `\textit{End-to-end ML cloud service}' \textit{(SI4)}: This option refers to use complete services made specifically for ML services by cloud companies. The operation mechanism is to load the model from model registry, which normally is in the same cloud services, this service configures endpoints for the models that user uploads, again without the need of manually build an API, and sends back the response with the prediction to the user. 
\end{itemize}

In addition to that decision, we identify other decisions as \textit{Transversal Decisions (TD)}  such as `\textit{Containerized}', `\textit{Model format}', `\textit{Request processing}', and `\textit{Communication protocol}'. While some of these decisions remain independent of the chosen serving infrastructure (e.g. `\textit{Containerized}'), it is worth noting that certain options from these decisions may lack compatibility with specific serving infrastructure (e.g. incompatibility between TensorFlow SavedModel format and TorchServe serving infrastructure). The Serving Infrastructure design decision determines others transversal decisions, such as the need of using a web framework to build an API, batching configurations, model format compatibility, communication protocols and integration with containerization tools such as docker. While these last decisions may optimize the interaction between users and the ML application, they are influenced by the underlying serving infrastructure. The identified transversal decisions are:
\begin{itemize}
    \item `\textit{Containerized}' \textit{(TD1)}: This option refers to the use of a containerization method, the most common is to use docker, but there are other methods such as using WebAssembly, that is a portable method employed in modern browsers. Also there is the option of avoiding a containerization layer.
    \item `\textit{Model format}' \textit{(TD2)}: This option refers to the format of the model used to serve them. Normally, the models are trained in common ML frameworks and then converted into a format that meets the requirements of the ML serving infrastructure.
    \item `\textit{Request processing}' \textit{(TD3)}: This option refers to the manner in which predictions are returned to the users, it can be real-time, returning the prediction as the user sends the request, or using a batch, returning the predictions after a number of requests are sent.
    \item `\textit{Communication protocol}' \textit{(TD4)}: This option refers to the protocol used for communication (e.g., REST, and gRPC) between user interface and the web server endpoint providing access to make inferences.
\end{itemize}

\begin{tcolorbox}
\textbf{Answer to RQ1:}
When serving models as a service, the main ML serving architectural design decision is the serving infrastructure, which refers to the architecture of \textit{(i)} `\textit{No runtime engine}', \textit{(ii)} `\textit{Runtime engine}', \textit{(iii)} `\textit{DL-specific software}' and \textit{(iv)} `\textit{End-to-end ML cloud service}'. Furthermore, there are additional transversal decisions that may be associated to the serving infrastructure such as \textit{(i)} `\textit{Containerized}', \textit{(ii)} `\textit{Model format}', \textit{(iii)} `\textit{Request processing}', and \textit{(iv)} `\textit{Communication protocol}'.
\end{tcolorbox}


\begin{table}[h]

    \caption{Design decisions and quality aspects from literature.}
    
    \label{tab:table_decisions}
    \scalebox{0.6}{
    
    \begin{tabular}{|cc|l|l|}
    
    \hline
    \multicolumn{2}{|c|}{\textbf{ML serving design decision}}                                                                                                                                                      & \multicolumn{1}{c|}{\textbf{Technological stack}}                                                                                                                                        & \multicolumn{1}{c|}{\textbf{Quality aspects}}                                                                                  \\ \hline
    \multicolumn{1}{|c|}{\multirow{4}{*}{\textbf{\begin{tabular}[c]{@{}c@{}}Serving\\ Infrastructure\\ (SI)\end{tabular}
    }}}                                          & \textbf{\begin{tabular}[c]{@{}c@{}}SI1: No runtime engine \\ and web framework\end{tabular}} & \begin{tabular}[c]{@{}l@{}}TensorFlow\cite{georgiou2022green,zhang2018,yao2021evaluating,koubaa2021cloud}, \\PyTorch\cite{georgiou2022green,zhang2018,yao2021evaluating}\end{tabular}                                                                                                                       & \begin{tabular}[c]{@{}l@{}}energy efficiency\cite{georgiou2022green,zhang2018},\\performance \\efficiency\cite{georgiou2022green,zhang2018,koubaa2021cloud},\\ analysability \\(function-level analysis)\cite{georgiou2022green}, \\ maintainability\\ (documentation)\cite{georgiou2022green}\end{tabular}                    \\ \cline{2-4} 
    \multicolumn{1}{|c|}{}                                                                                          & \textbf{\begin{tabular}[c]{@{}c@{}}SI2: Runtime engine\\ and web framework\end{tabular}}     & \begin{tabular}[c]{@{}l@{}}ONNX Runtime\cite{hampau2022empirical,zhang2020inferbench}, \\torch.jit \cite{zhang2020inferbench}, \\ TensorRT engine\cite{klimiato2022utilizing,koubaa2021cloud,yao2021evaluating, lenherr2021new}, \\ TFLite interpreter \cite{koubaa2021cloud,lenherr2021new}, \\ OpenVINO\cite{lenherr2021new}\end{tabular}                                                    & \begin{tabular}[c]{@{}l@{}}performance\\ efficiency\cite{hampau2022empirical,zhang2020inferbench},\\energy efficiency\cite{hampau2022empirical,yao2021evaluating,lenherr2021new}\end{tabular}                                                                                                           \\ \cline{2-4} 
    \multicolumn{1}{|c|}{}                                                                                          & \textbf{\begin{tabular}[c]{@{}c@{}}SI3: DL-specific \\ software\end{tabular}}                                                           & \begin{tabular}[c]{@{}l@{}}TF Serving\cite{klimiato2022utilizing,zhang2020inferbench}, \\ TorchServe\cite{klimiato2022utilizing,zhang2020inferbench}, \\ Triton Inference Server\cite{klimiato2022utilizing,zhang2020inferbench}\end{tabular}                                                                                                                               & \begin{tabular}[c]{@{}l@{}}performance\\ efficiency\cite{klimiato2022utilizing,zhang2020inferbench}\\ usability \\(ease-of-use) \cite{klimiato2022utilizing}\end{tabular}                  \\ \cline{2-4} 
    \multicolumn{1}{|c|}{}                                                                                          & \textbf{SI4: Cloud service}                                                                  & \begin{tabular}[c]{@{}l@{}}AWS Sagemaker\cite{klimiato2022utilizing}, \\ Vertex AI\cite{klimiato2022utilizing}, \\ Azure ML\cite{klimiato2022utilizing,escribano2023energy}, \\  Databricks\cite{klimiato2022utilizing,lwakatare2019taxonomy}\end{tabular} & \begin{tabular}[c]{@{}l@{}}usability\\ (ease-of-use)\cite{klimiato2022utilizing},\\ performance\\ efficiency\cite{escribano2023energy}\\ energy efficiency\cite{escribano2023energy},\\ scalability\cite{lwakatare2019taxonomy} \end{tabular} \\ \hline
    \multicolumn{1}{|c|}{\multirow{4}{*}{\textbf{\begin{tabular}[c]{@{}c@{}}Transversal\\ Decisions (TD)\end{tabular}}}} & \textbf{TD1: Containerized}                                                                       & \begin{tabular}[c]{@{}l@{}}Non-containerized\cite{hampau2022empirical}, \\ Docker\cite{hampau2022empirical,lwakatare2019taxonomy}, \\ WebAssembly\cite{hampau2022empirical}\end{tabular}                                                                                                                                       & \begin{tabular}[c]{@{}l@{}}portability\cite{hampau2022empirical}\\ energy efficiency\cite{hampau2022empirical},\\ performance\\ efficiency\cite{hampau2022empirical},\\ scalability\cite{lwakatare2019taxonomy}\end{tabular}              \\ \cline{2-4} 
    \multicolumn{1}{|c|}{}                                                                                          & \textbf{TD2: Model format}                                                                        & \begin{tabular}[c]{@{}l@{}}PyTorch \cite{klimiato2022utilizing,zhang2020inferbench}, \\ TF SavedModel \cite{klimiato2022utilizing,koubaa2021cloud,yao2021evaluating,zhang2020inferbench}, \\ ONNX \cite{koubaa2021cloud,yao2021evaluating,zhang2020inferbench}, \\ TorchScript\cite{zhang2020inferbench}, \\ TensorRT \cite{klimiato2022utilizing,koubaa2021cloud,yao2021evaluating,zhang2020inferbench,lenherr2021new}, \\ TF Lite \cite{koubaa2021cloud,lenherr2021new}, \\ aot or wasm \cite{hampau2022empirical}, \\ IR representation \cite{lenherr2021new}\end{tabular}                                                              & \begin{tabular}[c]{@{}l@{}}performance\\ efficiency \cite{koubaa2021cloud},\\interoperability \cite{klimiato2022utilizing,koubaa2021cloud,yao2021evaluating}\end{tabular}                                                                                                       \\ \cline{2-4} 
    \multicolumn{1}{|c|}{}                                                                                          & \textbf{\begin{tabular}[c]{@{}c@{}}TD3: Request \\ processing\end{tabular} }                                                                 & \begin{tabular}[c]{@{}l@{}}real-time \cite{kumara2022requirements,yao2021evaluating,yarally2023batching}, \\ batching\cite{kumara2022requirements,yao2021evaluating,yarally2023batching}\end{tabular}                                                                                                                                                          & \begin{tabular}[c]{@{}l@{}}performance\\ efficiency \cite{yao2021evaluating,yarally2023batching}\\ energy efficiency \cite{yarally2023batching,yao2021evaluating}\end{tabular}                                            \\ \cline{2-4} 
    \multicolumn{1}{|c|}{}                                                                                          & \textbf{\begin{tabular}[c]{@{}c@{}}TD4: Communication\\protocol\end{tabular}}                                                              & \begin{tabular}[c]{@{}l@{}}REST \cite{klimiato2022utilizing}, \\ gRPC \cite{klimiato2022utilizing,zhang2020inferbench}, \\ GraphQA \cite{klimiato2022utilizing}\end{tabular}                                                                                                                                                           & N/A                                                                                                                            \\ \hline
    \end{tabular}}
    
    \end{table}

\subsection{What are the  quality factors that literature have studied? (RQ2)}

 Table \ref{tab:table_decisions} presents the results to answer RQ2. From the collected results, we observe that in the majority of the decisions, at least one study considers performance efficiency, which primarily involves metrics such as runtime, latency, memory utilization, CPU utilization, or GPU utilization.

Regarding green AI characteristics, only in three out of the four serving infrastructures of Table \ref{tab:table_decisions}, researchers have considered to analyze the energy consumption or carbon emissions. Regarding the transversal decisions, only the containerization, and batching effect in the energy consumption has been considered.

The rest of the quality characteristics have also not been widely studied concerning ML serving ADDs; the majority of them are only considered with respect to one or two decisions. Moreover, there are no studies that have considered quality characteristics of the `\textit{Communication protocol}'.

\begin{tcolorbox}
\textbf{Answer to RQ2:}
The collected results highlight a predominant focus on investigating performance efficiency characteristics. Notably, there are few studies addressing the energy consumption aspect of ML serving components. There is limited research on how the consideration of available options from serving infrastructures and transversal decisions impacts the quality characteristics.

\end{tcolorbox}

\section{Threats to validity}
\textit{Study selection validity}: 
We consider this work as a preliminary study in our goal to achieve green ML serving. Three researchers were involved in the selection of papers. A validity threat may regard the selection of the seed papers. We considered seed papers from two important secondary peer-reviewed studies regarding the field of our work. To mitigate this thread, we used forward snowballing to expand the number of studies.

\textit{Data validity}: 
Regarding data validity, a threat is the data extraction bias and researcher bias. To mitigate this, data collection and data analysis were discussed among all the authors.

\textit{Research process validity}: 
Concerning the research process validity, ensuring the repeatability of our study is critical. We have documented the process to reproduce our work: the process is described in the Research Methodology section and the replication package is available online\footnote{Please refer to \url{https://zenodo.org/doi/10.5281/zenodo.10547444}}. Furthermore, our work is focused on Green ML serving. It is possible that our results may not be applicable to other fields. However, we intentionally limit the scope of this work to Green ML serving as it serves as a preliminary study for further research in this specific context.

\section{Discussion and outlook}


The primary ADD that emerges is the serving infrastructure, with four distinctive options: `\textit{No runtime engine}', `\textit{Runtime engine}', `\textit{DL-specific software}', and `\textit{End-to-end ML cloud service}'. Each of these choices presents different architecture and its own implications for ML serving efficiency. Transversal decisions, including `\textit{Containerized}', `\textit{Model format}', `\textit{Request processing}', and `\textit{Communication protocol}', were identified, showcasing their dependence on the selected serving infrastructure.

In our exploration of the serving infrastructure decision in ML serving, a comprehensive understanding of the associated advantages and drawbacks is key. Opting for `\textit{No runtime engine}' can minimize resource overhead but may sacrifice optimization capabilities and the need to build an API. Conversely, employing a `\textit{Runtime engine}' enhances model optimization but introduces complexity, again with the need to build an API. `\textit{DL-specific software}' offers specialized optimizations for ML frameworks and eliminates the need for building an API but may lack versatility. When considering an `\textit{End-to-end ML cloud service}', ML serving is simplified by cloud services, it provides a ready-to-use alternative and has integration with the entire ML services provided by the same cloud company but introduces dependencies and privacy concerns.

Our results have helped us to identify the quality characteristics that are currently addressed by researchers and practitioners in the field. According to the results, performance efficiency is the most considered quality characteristic  in literature. Notably, energy efficiency emerged as a less-explored dimension, with only a handful of studies delving into the environmental impact of ML serving components. This indicates that practitioners should also employ additional metrics to evaluate their ML serving architecture, including energy efficiency, ease of use, scalability, or interoperability metrics, in addition to performance efficiency metrics.

Remarkably, our analysis revealed a research gap in the current literature, as we did not find any study addressing the relationship between quality aspects other than performance efficiency, and the environmental impact of ML serving systems. This emphasizes the need for a more comprehensive understanding of the various quality factors influencing the energy footprint of ML serving (e.g., usability, reliability, portability, etc.).

Regarding the serving infrastructure decision, which is considered to be the main design decision in our study, from the studies that address energy consumption of serving infrastructure options, only three of them compare options from serving infrastructures \cite{hampau2022empirical, georgiou2022green, yao2021evaluating}. Georgiou \etal \cite{georgiou2022green} is the only work focused on one serving infrastructure (\textit{SI1}) making inferences directly with the ML framework. This indicates a clear lack of research on energy consumption of ML serving infrastructures, and ML serving itself. Hampau \etal \cite{hampau2022empirical}, considered ``AI containerization strategies'' to compare a combination of two decisions considered in our results: \textit{SI2} and \textit{TD1}. Studying the options within specific \textit{SIs} may be more significant and is a research opportunity. For instance, a comparative analysis of various options within runtime engines (e.g., ONNX runtime, OpenVINO and torch.jit) based on the discussed quality characteristics can empower practitioners to improve the efficiency of ML serving and make more informed decisions aligned with their application requirements.

This preliminary study lays the foundation for our future work, identifying the ADDs that practitioners should take into consideration when building green ML serving systems and the quality characteristics that should be studied regarding ML serving ADDs. Conducting in-depth analysis of the energy efficiency of each serving infrastructure option may provide practitioners with clearer guidelines on building green ML serving systems. Additionally, it would be valuable to study other less common quality characteristics such as ease-of-use, scalability or interoperability with respect to the presented ADDs in this study.

\section*{Acknowledgment}
This paper has been funded by the GAISSA Spanish research project (ref. TED2021-130923B-I00; MCIN/AEI/10.13039/501100011033).

\clearpage
\bibliographystyle{ACM-Reference-Format}
\bibliography{references}

\end{document}